\documentclass[letterpaper]{article} 
\usepackage{aaai25}  
\usepackage{times}  
\usepackage{helvet}  
\usepackage{courier}  
\usepackage[hyphens]{url}  
\usepackage{graphicx} 
\usepackage{bm}
\usepackage{epsfig}
\usepackage{amsmath}
\usepackage{amssymb}
\usepackage{booktabs}
\usepackage{multirow}
\usepackage{xcolor}
\usepackage{natbib}  
\usepackage{caption} 
\usepackage{algorithm}
\usepackage{algorithmic}
\usepackage[T1]{fontenc}

\usepackage{newfloat}
\usepackage{listings}
\DeclareCaptionStyle{ruled}{labelfont=normalfont,labelsep=colon,strut=off} 
\floatstyle{ruled}
\newfloat{listing}{tb}{lst}{}
\floatname{listing}{Listing}
\title{Navigating Label Ambiguity for Facial Expression Recognition in the Wild}
\author {
    JunGyu Lee\textsuperscript{\rm 1,2}\equalcontrib, 
    Yeji Choi\textsuperscript{\rm 1,3}\equalcontrib,
    Haksub Kim\textsuperscript{\rm 1},
    Ig-Jae Kim\textsuperscript{\rm 1,2},
    Gi Pyo Nam\textsuperscript{\rm 1,2}
}
\affiliations {
    \textsuperscript{\rm 1} Korea Institute of Science and Technology, Seoul, Korea\\
    \textsuperscript{\rm 2} AI-Robotics, KIST School, University of Science and Technology, Daejeon, Korea\\
    \textsuperscript{\rm 3} Yonsei University, Seoul, Korea\\
    \{jungyu0413, cyjcyj91, hskim, drjay, gpnam\}@kist.re.kr\\
}

\usepackage{bibentry}

\begin{document}

\maketitle

\begin{abstract}
Facial expression recognition (FER) remains a challenging task due to label ambiguity caused by the subjective nature of facial expressions and noisy samples. Additionally, class imbalance, which is common in real-world datasets, further complicates FER. Although many studies have shown impressive improvements, they typically address only one of these issues, leading to suboptimal results. To tackle both challenges simultaneously, we propose a novel framework called Navigating Label Ambiguity (NLA), which is robust under real-world conditions. The motivation behind NLA is that dynamically estimating and emphasizing ambiguous samples at each iteration helps mitigate noise and class imbalance by reducing the model's bias toward majority classes. To achieve this, NLA consists of two main components: Noise-aware Adaptive Weighting (NAW) and consistency regularization. Specifically, NAW adaptively assigns higher importance to ambiguous samples and lower importance to noisy ones, based on the correlation between the intermediate prediction scores for the ground truth and the nearest negative. Moreover, we incorporate a regularization term to ensure consistent latent distributions. Consequently, NLA enables the model to progressively focus on more challenging ambiguous samples, which primarily belong to the minority class, in the later stages of training. Extensive experiments demonstrate that NLA outperforms existing methods in both overall and mean accuracy, confirming its robustness against noise and class imbalance. To the best of our knowledge, this is the first framework to address both problems simultaneously.
\end{abstract}

\section{Introduction}

Facial expressions play a crucial role in non-verbal communication in daily life. The growing use of remote communication through video connections has heightened the importance of facial expression recognition (FER) technology.
It has many valuable applications ranging from human-computer interaction in customer service to remote mental health care support. In recent years, owing to the emergence of large-scale datasets collected in the wild, such as FERPlus~\cite{barsoum2016training_fer+}, RAF-DB~\cite{li2017reliable_rafdb}, and AffectNet~\cite{mollahosseini2017affectnet}, many deep learning-based FER methods~\cite{wang2020region_RAN, zhao2021robust_LDL, li2021adaptively_KTN, xue2021transfer, xue2022vision_APViT, zheng2023poster, mao2023poster++} have been developed rapidly and achieved impressive performance.

\begin{figure}[t]
    \centerline{\includegraphics[width=1\linewidth]{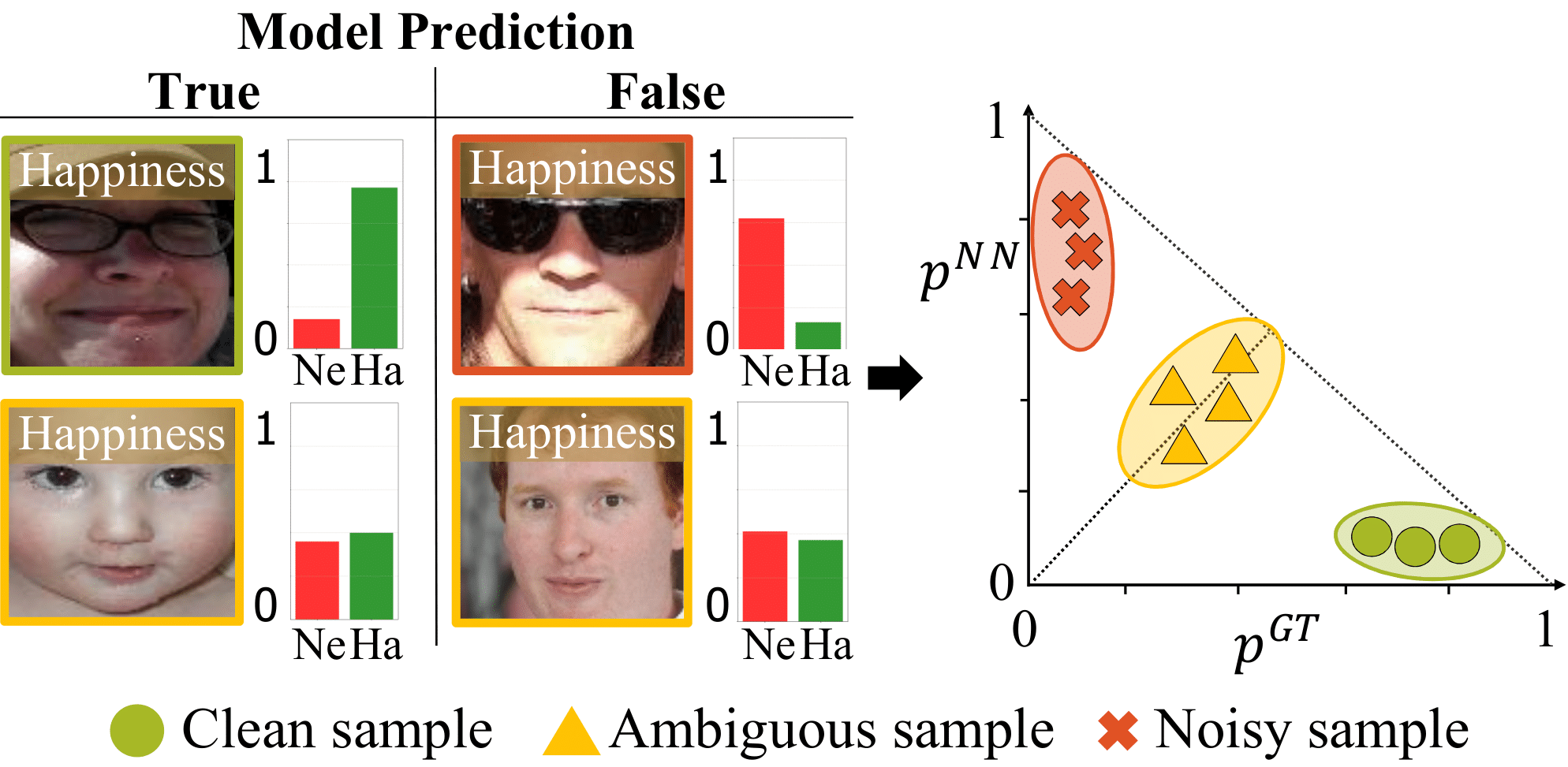}} 
    \caption{\textbf{Visualization of estimating sample ambiguity.}  
    The visual analysis on the right illustrates how the correlation between the prediction scores for ground truth (GT) and nearest negative (NN) serves as a criterion for categorizing samples as clean, ambiguous, or noisy. The prediction scores for the GT and the NN are represented by green and red bars in the probability distribution on the left side, respectively (Ne: Neutral, Ha: Happiness).}
    \label{fig:motivation} 
\end{figure}

One of the main challenges in FER is label ambiguity, which arises from variability in the expression and interpretation of facial emotions across individuals. This complexity makes it difficult to reliably categorize facial expressions into one of the seven basic classes (e.g., neutral, happiness, fear, etc.). Several approaches~\cite{she2021dive_DMUE, zhang2021relative_RUL, zhang2022man} have attempted to mitigate this ambiguity by comparing pairwise samples or aligning predictions from multiple branches. Meanwhile, class imbalance presents another critical issue in FER, where certain expressions are overrepresented in datasets (e.g., happiness and neutral), while others (e.g., fear and disgust) are underrepresented. MEK~\cite{zhang2024leave_MEK} has been proposed to address this imbalance by emphasizing minority classes through re-balanced attention consistency. However, existing methods focus on either label ambiguity or class imbalance in isolation, resulting in suboptimal performance. 
Moreover, studies on label ambiguity often take a narrow approach, focusing solely on noisy samples without leveraging their full potential to address class imbalance.

To tackle these challenges together, we prioritize sample ambiguity as a key criterion.
Drawing inspiration from dynamic weight adjustment, we assign different weights to each sample based on its estimated ambiguity. 
Fig.~\ref{fig:motivation} illustrates the process of estimating sample ambiguity by analyzing the correlation between the prediction scores for the ground truth (GT) and the nearest negative (NN)—the class with the highest prediction score excluding the GT. 
When the model's prediction is true, it generates a GT score much higher than the NN score for clean samples, whereas for ambiguous samples, the scores are similar. 
In contrast, when the model's prediction is false, it produces the opposite result.
Based on these discrepancies, we categorize the samples by facial expression class, observing that most of the ambiguous and noisy samples are concentrated in the minority class. This observation motivates us to leverage this relationship to emphasize the minority class during model training.

Building on these insights, we propose a new framework called Navigating Label Ambiguity (NLA). The core idea of NLA is to adaptively assign weights based on the estimated sample ambiguity at each training step. This strategy allows the model to progressively focus on ambiguous samples in the minority class as training progresses, while penalizing noisy samples.
To achieve this, NLA comprises two main components: Noise-aware Adaptive Weighting (NAW) and consistency regularization. Specifically, NAW assigns higher weights to ambiguous samples and lower weights to noisy ones by analyzing the correlation between the prediction scores of the ground truth and the nearest negative.
In particular, we apply different forms of multivariate Gaussian kernels for NAW, based on intermediate prediction results and the training epoch, ensuring that ambiguous samples within the minority class receive increased attention after the learning for the majority class has saturated.
Additionally, consistency regularization further enhances the reliability of NLA by aligning the latent distributions of the original and flipped images using Jensen-Shannon Divergence.

The main contributions of our method are as follows:
\begin{itemize}
\item We propose a novel framework called Navigating Label Ambiguity (NLA) by leveraging adaptive weighting and consistency regularization. To the best of our knowledge, this is the first attempt to address both class imbalance and noisy labels by handling label ambiguity. 
\item We introduce a Noise-aware Adaptive Weighting (NAW) that dynamically assigns weights based on sample ambiguity, allowing the model to focus progressively on ambiguous samples while minimizing the impact of noise.
\item NLA demonstrates superior performance in overall and mean accuracy on in-the-wild datasets. Additionally, extensive experiments under different noise and class imbalance conditions confirm the robustness of our method.

\end{itemize}

\section{Related Work}
\subsection{Facial Expression Recognition}
In-the-wild FER scenarios present two major challenges: 1) noisy labels, which are common in image classification but more prevalent in FER due to the inherent ambiguity of facial expressions, and 2) class imbalance resulting from varying expression frequencies. 
Recently, numerous methods have been proposed to alleviate these issues. 
For instance, SCN~\cite{wang2020suppressing_SCN} uses re-labeling and ranking regularization to mitigate noise, while EAC~\cite{zhang2022learn_EAC} minimizes overfitting to noisy samples by aligning Class Activation Maps~\cite{zhou2016learning_CAM} of the original and flipped images.
LA-Net \cite{wu2023net_LA-Net} utilizes landmark data to enhance expression features via expression-landmark interactions.  
Alternatively, paying more attention to ambiguity, DMUE \cite{she2021dive_DMUE} explores latent distributions with multiple branches based on uncertainty estimation, and RUL~\cite{zhang2021relative_RUL} uses a multi-branch framework with feature mixup to learn from relative sample difficulty. MAN~\cite{zhang2022man} employs a two-branch network with a co-division module to enhance discriminative ability by focusing on clean samples.

Regarding class imbalance, Face2Exp~\cite{zeng2022face2exp} improves FER by using large-scale unlabeled face recognition data within a meta-optimization framework, while MEK~\cite{zhang2024leave_MEK} emphasizes minority classes through re-balanced attention consistency and label smoothing. Despite these advances, real-world FER is still limited to improving generalization due to the combined challenges of noisy labels and class imbalance. Unlike previous works, our method addresses both issues simultaneously by dynamically exploring each sample's ambiguity.

\subsection{Learning with Ambiguity}
Recent approaches to handling ambiguity can be broadly categorized into small loss selection and disagreement/agreement strategies. Previous research~\cite{arpit2017closer_small_loss} has shown that deep neural networks initially learn simple patterns, leading to the assumption that small-loss samples are treated as clean. Inspired by this, MentorNet~\cite{jiang2018mentornet} uses a teacher-student framework, where the student network is trained only on small-loss samples selected by the teacher network. Co-teaching~\cite{han2018co_Co-teaching} further cross-updates two networks by exchanging small-loss samples in each mini-batch.

Alternatively, the disagreement/agreement strategy focuses on instances where predictions between two networks differ, which is similar to hard sample mining. 
As representative works, Decoupling~\cite{malach2017decoupling} and Co-teaching+~\cite{yu2019does_coteaching+} update networks based on these disagreements. In contrast, JoCoR~\cite{wei2020combating_JoCoR} aligns the predictions of the two networks using Jensen-Shannon divergence to ensure agreement. Drawing ideas from this approach, we apply a consistency regularization method that aligns the probabilities between the original data and its flipped version within a single network.

\begin{figure*}[htb!]
    \centerline{\includegraphics[width=1\textwidth]{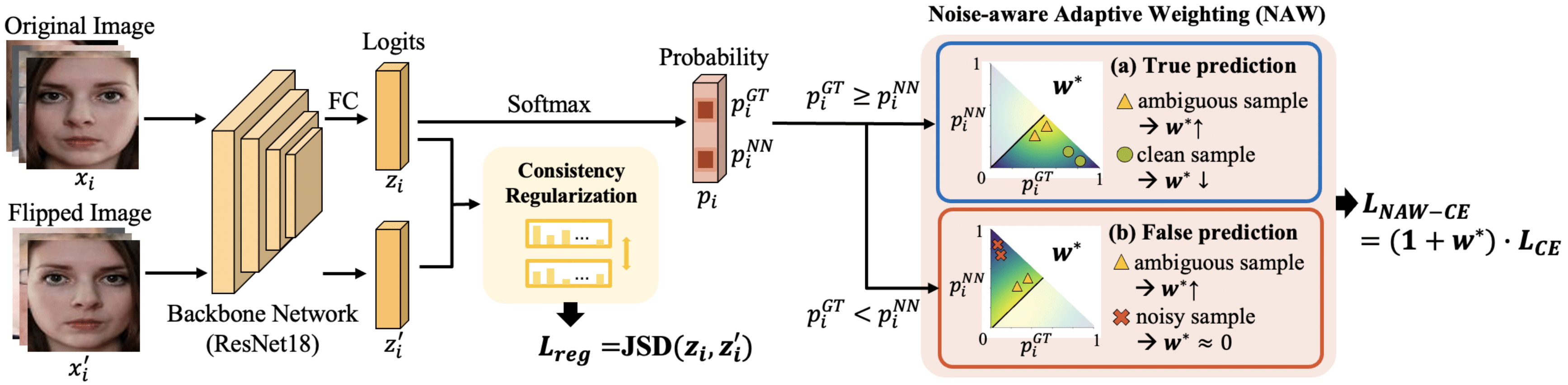}}
    \caption{\textbf{The framework of Navigating Label Ambiguity (NLA).} 
    NLA consists of two main components: 1) a Noise-aware Adaptive Weighting (NAW), which dynamically assigns weights to each sample based on the intermediate prediction scores for GT and NN, and 2) consistency regularization using pairs of original and horizontally flipped images.}
    \label{fig:main}
\end{figure*}

\subsection{Learning with Imbalanced Data}
Training samples in wild datasets often exhibit imbalanced class distributions, leading to models biased toward majority classes. One basic approach to address this issue is re-balancing class distributions using different sampling frequencies for each class. Decoupling~\cite{kang2019decoupling_imbalanced} evaluates various sampling strategies and finds that progressively balanced sampling is particularly effective. Dynamic Curriculum Learning (DCL)~\cite{wang2019dynamic_DCL} introduces an adaptive sampling strategy that starts with random sampling and later shifts focus to minority classes.

Another approach involves re-weighting the loss for each class to ensure balanced contributions. CB loss~\cite{cui2019class_CB} uses the effective number, which is inversely proportional to class size, ensuring equal loss contribution. Balanced Softmax~\cite{ren2020balanced} adjusts predicted logits based on label frequencies by considering class priors before calculating the final loss. Our approach combines these strategies by proposing loss re-weighting based on sample ambiguity, rather than class distribution, allowing the model to prioritize ambiguous samples in the minority class during later training stages.

\section{Method}
In this section, we provide a brief review of the preliminaries related to our work. We then introduce our proposed method, Navigating Label Ambiguity (NLA), which comprises two main components: a noise-aware adaptive weighting and consistency regularization. Finally, we present the overall training objectives for our networks.
The framework of our proposed method is illustrated in Fig.~\ref{fig:main}.

\subsection{Preliminaries}
Given a $K$-class, $N$-sample FER dataset $\mathcal{D} = \{(x_i, y_i)\}_{i=1}^{N}$, where $x_i$ denotes the $i$-th face image and $y_i \in \{1, \cdots, K\}$ represents its corresponding ground-truth label, respectively.
Generally, a feature extractor based on a deep neural network utilizes a fully connected layer to predict the logit $z_i$ for a given $x_i$ and calculates the probability of the $k$-th class using a softmax function defined as: 
\begin{equation}
    p_i^k = \frac{e^{z_{ik}}}{\sum_{j=1}^{K} e^{z_{ij}}},
\end{equation}
where $z_{ik}$ and $z_{ij}$ represent the output logits for classes $k$ and $j$, respectively. 
Then, most multi-class classification models are optimized by minimizing the cross-entropy loss, defined as follows:
\begin{equation}
    \mathcal{L}_{\text{CE}}= -\frac{1}{N} \sum_{i=1}^{N} \log\,p_i^{y_i}.
\end{equation}

However, cross-entropy loss often produces suboptimal results in noisy and imbalanced datasets because it treats all samples with equal importance~\cite{ghosh2017robust}. It also tends to be biased towards easy samples, which usually belong to majority classes, weakening gradients for minority classes~\cite{lin2017focal}.

\subsection{Noise-aware Adaptive Weighting (NAW)}
Our goal is to adaptively assign weights to the loss function for each sample based on their ambiguity within the training pipeline. 
Following the motivation described in Fig.~\ref{fig:motivation}, the sample ambiguity can be estimated by considering the correlation between the prediction scores for the ground truth (GT) and the nearest negative (NN). 
These prediction scores for the GT and the NN are defined as:
\begin{equation}
    p_i^{GT} = p_i^{y_i}, \quad \quad p_i^{NN} = \underset{k \neq y_i}{\mathrm{max}}\;p_i^k.    
\end{equation}

Building on this, we introduce the noise-aware adaptive 
weighting (NAW) through a multivariate
Gaussian kernel 
that takes these two prediction scores as inputs, with a predetermined mean vector $\bm{\mu}$ and covariance matrix $\bm{\Sigma}$, formulated as follows:
\begin{equation}
    w^{*}(\mathbf{p}_i| \bm{\mu}, \bm{\Sigma})= C\cdot\exp \left( -\frac{1}{2} (\mathbf{p}_i - \bm{\mu})^T \bm\Sigma^{-1} (\mathbf{p}_i - \bm{\mu}) \right),
\end{equation}
where
\[
\mathbf{p}_i = \begin{bmatrix}
p_i^{GT} \\
p_i^{NN}
\end{bmatrix},
\quad
\bm{\mu} = \begin{bmatrix}
\mu_1 \\
\mu_2
\end{bmatrix},
\quad
\bm{\Sigma} = \begin{bmatrix}
\sigma_{11} & \sigma_{12} \\
\sigma_{21} & \sigma_{22}
\end{bmatrix},
\]
and $C=(2\pi \sqrt{|\bm{\Sigma}|})^{-1}$ is the normalizing constant. For convenience of presentation, we let $w^{*}(\mathbf{p}_i| \bm{\mu}, \bm{\Sigma})=w^{*}(\mathbf{p}_i)$. 

Then, using the above weights, we can define a NAW-based cross-entropy loss, $\mathcal{L}_\text{NAW-CE}$, as follows:
\begin{equation}
\mathcal{L}_{\text{NAW-CE}} = (1+w^{*}(\mathbf{p}_i)) \cdot \mathcal{L}_{CE}.
\end{equation}

\begin{figure}[t]
    \centerline{\includegraphics[width=1\columnwidth]{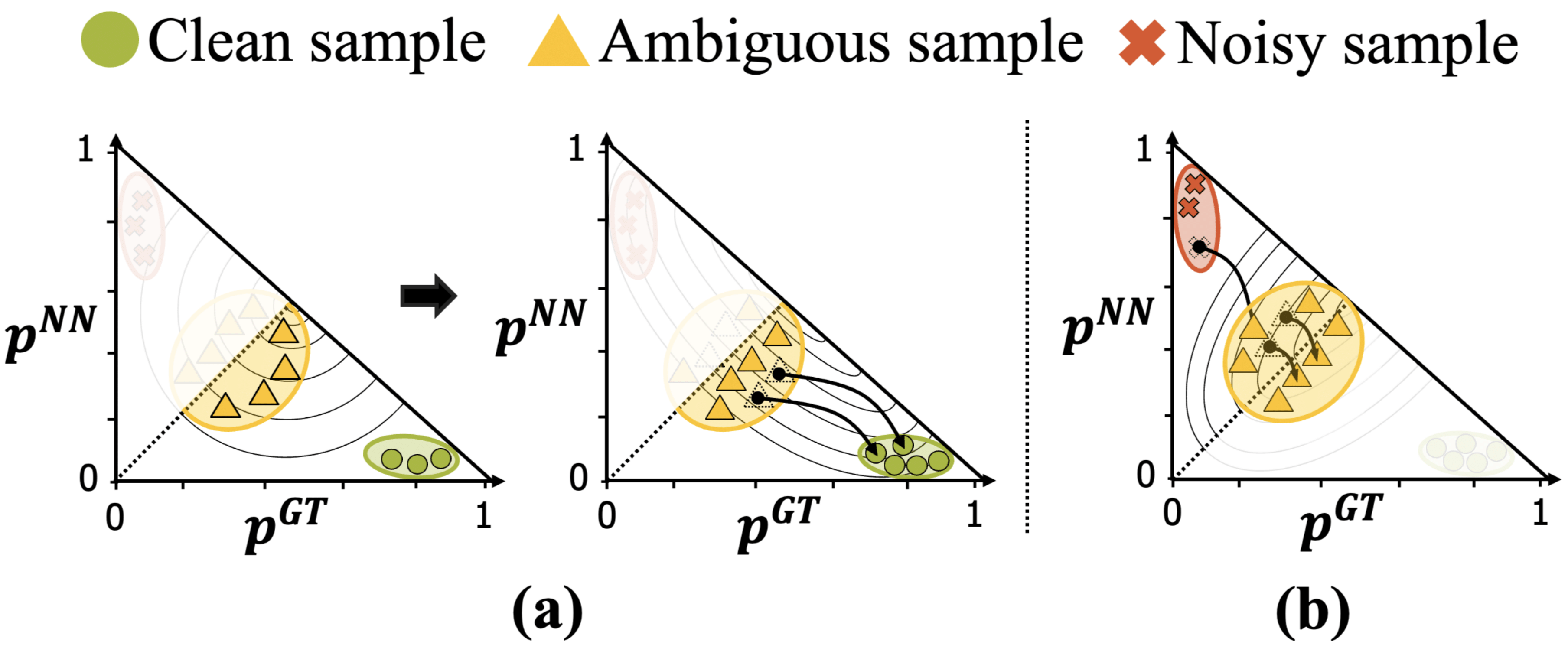}}
    \caption{\textbf{Visualization of the effect of NAW by prediction results.} 
    This figure illustrates how NAW enhances the model's ability to distinguish between clean, ambiguous, and noisy samples throughout the training process. (a) shows the results when the prediction is true, and (b) shows the results when the prediction is false.
    }
    \label{fig:NAW}
\end{figure}

Geometrically, the mean vector determines the center of the Gaussian kernel, where the weight is maximized, and the covariance matrix controls the shape of its contour. 
For example, if the covariance matrix is an identity matrix, the contours are isotropic (circular), resulting in weights that decrease at the same rate in all directions. If not, the contours take an elliptical shape, with weights decreasing gradually along the major axis and more sharply along the minor axis.
Leveraging these properties, as shown in Fig.~\ref{fig:NAW}, we design two different forms of NAW with distinct mean vectors and covariance matrices, depending on whether the intermediate prediction is true or false, as follows.

\paragraph{When the prediction is true.} 
In this case, the samples can be considered as either a clean sample (where $p_i^{GT}$ is much higher than $p_i^{NN}$), or an ambiguous sample (where $p_i^{GT}$ and $p_i^{NN}$ are both close to 0.5). 
To ensure that the ambiguous samples receive the highest weight, while the relatively easy ones are assigned lower weights, we set the mean vectors in the true case, $\bm{\mu}_t=[0.5,0.5]^T$.
As training progresses, the model becomes more discriminative on the ambiguous samples, resulting in more samples being classified as clean.
At this point, if an isotropic Gaussian kernel is applied throughout the entire learning process, information about clean samples may be lost due to their lower emphasis. 
To prevent this situation, we use a covariance scheduler (CS) that is exponentially adjusted at each epoch, modifying the contour from isotropic to an elongated ellipse along the $y=-x$ line. Fig.~\ref{fig:NAW} (a) visually depicts this process. 
The covariance matrix in the true case is defined as follows:
\[
\bm{\Sigma}_t = \begin{bmatrix} \sigma_{11} & \text{CS} \cdot \sigma_{12} \\ \text{CS} \cdot \sigma_{21} & \sigma_{22} \end{bmatrix},
\]
with
\begin{equation}
\text{CS}(e, E) = 
1 - \exp\left(\frac{\text{-10} \cdot e}{E}\right),
\end{equation}
where $e$ represents the current training epoch and $E$ denotes the total number of epochs.

\paragraph{When the prediction is false.} 
In this case, the samples can be considered as either a noisy sample (where $p_i^{GT}$ is much lower than $p_i^{NN}$), or an ambiguous sample (where $p_i^{GT}$ and $p_i^{NN}$ are similar). 
Since the ambiguous samples in the false case have been observed to have a lower $p_i^{NN}$ compared to those in the true case, we set the mean vector in the false case, $\bm{\mu}_f=[0.3,0.15]^T$. 
Additionally, to minimize the impact of noise, we experimentally set the covariance matrix, $\bm{\Sigma}_f$, so that its contour forms an elliptical shape elongated along the $y=x$ axis.
As shown in Fig.~\ref{fig:NAW} (b), NAW enhances the model's ability to distinguish ambiguous samples that were previously misclassified, while gradually increasing the weight of samples that were initially categorized as noise but are beneficial for learning. 
As training progresses to the later stages, most of the remaining ambiguous samples in the false cases belong to minority classes. NAW pays increased attention to these samples, improving the model's discriminative ability for minority classes.

\subsection{Consistency Regularization}
Consistency regularization is one of the effective way to enhance a model's reliability in the presence of label ambiguity. 
Inspired by this, we apply a regularization technique to ensure consistent predictions across different views of the same sample.
This is particularly essential for our method, as NAW assigns the weights based on the prediction scores for each sample at each epoch.
To achieve this, we adopt Jensen-Shannon Divergence (JSD), which measures the alignment of two distributions, using output logits of the original image, $x_i$, and its horizontally flipped image, $x'_i$.
Let $z_i$ and $z'_i$ be the output logits from $x_i$ and $x_i^{\prime}$, respectively. 
Then, the consistency regularization loss can be defined as follows:
\begin{equation}
\mathcal{L}_{\text{reg}} = D_{KL}(z_i \parallel \frac{z_i + z'_i}{2}) + D_{KL}(z_i' \parallel \frac{z_i + z'_i}{2}),
\end{equation}
where $D_{KL}(\cdot||\cdot)$ denotes the Kullback–Leibler (KL) divergence.

\subsection{Loss Functions}
In summary, the final loss function of our network is defined as follows:
\begin{equation}
\mathcal{L}_{\text{total}} = \lambda \cdot \mathcal{L}_{\text{NAW-CE}} + (1 - \lambda) \cdot \mathcal{L}_{\text{reg}},
\end{equation}
where $\lambda$ is a weighting factor. 

\begin{table*}[t!]
\centering
\resizebox{\linewidth}{!}{
\begin{tabular}{cccccccccccc}
\toprule
\textbf{Method} & \textbf{Conference} & \textbf{Overall} & \textbf{Mean} & {Happiness} & {Neutral} & {Sadness} & {Surprise} & {Disgust} & {Anger} & {Fear} \\ 
\midrule
Baseline & - & 87.42 & 78.53 & 95.44 & 88.53 & 85.56 & 83.59 & 58.75 & 78.40 & 59.46 \\ 
CB & CVPR 19 & 88.04 & 79.26 & 95.11 & \underline{90.74} & 84.73 & 86.93 & 64.38 & 73.46 & 59.46 \\ 
BBN & CVPR 20 & 87.39 & 78.19 & 94.59 & \textbf{91.62} & 84.94 & 84.80 & 61.88 & 77.78 & 52.70 \\ 
RUL & NeurIPS 21 & 88.66 & 81.66 & \underline{95.78} & 87.06 & 86.19 & \textbf{89.36} & 65.00 & \underline{83.33} & 64.86 \\ 
EAC & ECCV 22 & 89.05 & 81.09 & 95.27 & 88.97 & \underline{90.17} & \underline{87.84} & 61.25 & \underline{83.33} & 60.81 \\ 
MEK & NeurIPS 23 & \underline{89.77} & \underline{82.44} & \textbf{96.37} & 89.56 & 89.33 & \underline{87.84} & \underline{66.89} & 80.86 & \underline{66.22} \\
\textbf{NLA(Ours)} & - & \textbf{89.93} & \textbf{83.87} & 95.70 & 88.97 & \textbf{90.38} & 87.23 & \textbf{70.00} & \textbf{84.57} & \textbf{70.27} \\ 
\bottomrule
\end{tabular}}
\caption{\textbf{Comparison with other methods on RAF-DB using pre-trained ResNet-18 as backbone.} We achieve significant improvement in minority classes (e.g., Disgust, Anger, and Fear), leading to the best performance in both overall and mean accuracy (\textbf{Bold}: best, \underline{underline}: second best).}
\label{tab:rafdb}
\end{table*}

\begin{table*}[t]
\centering
\resizebox{\linewidth}{!}{ 
\begin{tabular}{cccccccccc}
\toprule
\textbf{Method} & \textbf{Conference} & \textbf{Overall(=Mean)} & {Happiness} & {Neutral} & {Sadness} & {Surprise} & {Disgust} & {Anger} & {Fear} \\ 
\midrule
BBN  & CVPR 20 & 60.76 & 87.00 & 57.10 & \textbf{66.80} & 54.90 & 30.10 & 58.30 & \textbf{71.10} \\ 
RUL  & NeurIPS 21 & 61.56 & \underline{90.50} & 62.40 & 64.70 & 60.80 & 34.20 & \textbf{69.30} & 49.00 \\ 
EAC  & ICCV 22 & 65.17  & \textbf{91.40} & 64.50 & \underline{65.70} & \underline{61.60} & 45.80 & 66.30 & 60.90 \\ 
MEK & NeurIPS 23 & \underline{65.73} & 86.20 & 59.00 & 64.20 & 57.80 & \textbf{61.90} & \underline{66.50} & 64.50 \\ 
\textbf{NLA(Ours)} & - & \textbf{67.06} & 88.60 & \textbf{65.60} & 63.60 & \textbf{64.20} & \underline{61.20} & 60.40 & \underline{65.80} \\ 
\bottomrule
\end{tabular}}
\caption{\textbf{Comparison with other methods on AffectNet using pre-trained ResNet-18 as backbone.} Our method achieves over 60\% prediction accuracy across all classes, yielding the best overall accuracy (\textbf{Bold}: best, \underline{underline}: second best).}
\label{tab:affectnet}
\end{table*}

\begin{figure}[t]
    \centerline{\includegraphics[width=1\linewidth]{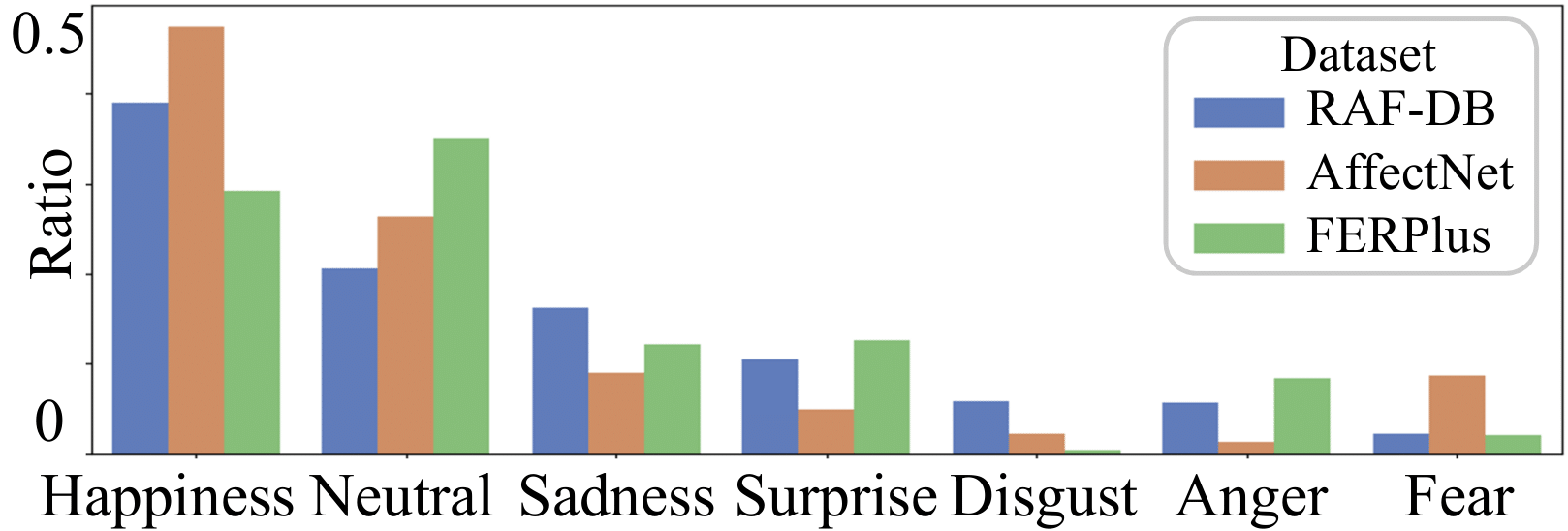}}
    \caption{Imbalanced distribution of training samples in the wild FER dataset.}
    \label{fig:dataset_distribution} 
\end{figure}

\section{Experiments}
In this section, we evaluate the effectiveness of NLA on three in-the-wild FER benchmarks, considering variations in label noise and class imbalance. Additionally, we verify the contribution of each component of NLA through comprehensive ablation studies and provide visualization analysis.

\subsection{Experimental Settings}
\paragraph{Datasets.}
\textbf{RAF-DB}~\cite{li2017reliable_rafdb} contains 30,000 face images labeled with 7 basic and compound expressions by 40 trained annotators. We use a subset with 7 basic expressions, comprising 12,271 training images and 3,068 test images. 
\textbf{FERPlus}~\cite{barsoum2016training_fer+}, an extension of FER2013~\cite{goodfellow2013challenges_fer2013}, provides 8 expression labels with the addition of Contempt, labeled by 10 annotators. For a fair comparison, we use the same 7 basic classes, totaling 28,709 training images and 3,589 test images. 
\textbf{AffectNet}~\cite{mollahosseini2017affectnet}, the largest FER dataset, contains 450,000 face images with 7 basic expressions and a contempt label. In our experiments, we use only the 7 basic classes, which include 283,901 training images and 3,500 test images.
As shown in Fig.~\ref{fig:dataset_distribution}, the training samples in all datasets are notably imbalanced.

\paragraph{Implementation Details.}
We utilize ResNet-18~\cite{he2016deep_resnet18} pre-trained on MS-Celeb-1M~\cite{guo2016ms_msceleb}, following previous works~\cite{zhang2022learn_EAC, wu2023net_LA-Net}, for a fair comparison. 
All face images are aligned and cropped based on three landmarks~\cite{wang2019adaptive_crop}, and then resized to 224$\times$224. 
For network training, we use the Adam optimizer~\cite{kingma2014adam} with a weight decay of 0.0001 and employ the ExponentialLR scheduler~\cite{li2019exponential} with a gamma value of 0.9 and an initial learning rate of 0.0001. 
We set $\lambda$ to 0.5, the batch size to 32, and the maximum training epoch to 60, with the best performance observed at epoch 40. 
We determine each $\Sigma$ based on the diagonal element $\sigma_{11} = 0.8$ in both cases, with the ratio of the major to minor axis being 2:1 and 6:1 for the true and false cases, respectively.
All experiments are conducted on a single NVIDIA A100.

\subsection{Comparison with Existing Methods}
We conduct evaluations on the RAF-DB and AffectNet benchmarks, comparing the performance of our proposed method with existing FER methods that use pre-trained ResNet-18 as the backbone, including CB~\cite{cui2019class_CB}, BBN~\cite{zhou2020bbn}, KTN~\cite{li2021adaptively_KTN}, RUL~\cite{zhang2021relative_RUL}, EAC~\cite{zhang2022learn_EAC}, MEK~\cite{zhang2024leave_MEK}, and LA-Net~\cite{wu2023net_LA-Net}. 
As shown in Table~\ref{tab:rafdb}, NLA outperforms other methods on RAF-DB, achieving an overall accuracy of 89.93\% and a mean accuracy of 83.87\%. 
Notably, while several methods, including RUL and EAC, have steadily improved overall accuracy by addressing noisy labels, their improvements in mean accuracy are relatively limited.
A closer examination reveals that these improvements are primarily driven by enhanced performance in majority classes such as `Happiness' and `Neutral', while performance in minority classes such as `Fear' and `Disgust' remains suboptimal. 
This bias highlights the need for approaches that address both noisy labels and class imbalance in in-the-wild datasets.
In contrast, NLA surpasses the current state-of-the-art method, MEK, by approximately 4\% 
in minority classes and is the first to achieve over 70\% accuracy in these classes, demonstrating its robustness in addressing class imbalance.

Table~\ref{tab:affectnet} presents the performance comparison on AffectNet, where the test set is class-balanced, making overall accuracy equal to mean accuracy. 
Similar to the previous results, RUL and EAC achieve high accuracy in the majority class, `Happiness', but still suffer from performance degradation in the minority classes.
Although MEK improves accuracy for minority classes through re-balancing techniques, it sacrifices accuracy in majority classes, `Happiness' and `Neutral'.
Conversely, NLA maintains high accuracy in the majority class while achieving over 60\% accuracy across all classes, leading to the best overall accuracy.

\begin{table}[t!]
\centering
\resizebox{\linewidth}{!}{
\begin{tabular}{cccccc}
\toprule
\textbf{Method} & \textbf{Noise(\%)} & \textbf{RAF-DB} & \textbf{AffectNet} & \textbf{FERPlus} \\ \midrule
Baseline & 10 & 81.01 & 57.24 & 83.29 \\ 
SCN & 10 & 82.18 & 58.58 & 84.28 \\ 
RUL & 10 & 86.17 & 60.54 & 86.93 \\ 
EAC & 10 & 88.02 & 61.11 & 87.03 \\ 
LA-Net & 10 & 88.75 & 62.85 & 88.02 \\ 
\textbf{NLA(Ours)} & 10 & \textbf{88.83$\pm$0.11} & \textbf{63.52$\pm$0.08} & \textbf{88.20$\pm$0.07} \\ 
\midrule
Baseline & 20 & 77.98 & 55.89 & 82.34 \\ 
SCN & 20 & 80.10 & 57.25 & 83.17 \\ 
RUL & 20 & 84.32 & 59.01 & 85.05 \\ 
EAC & 20 & 86.05 & 60.29 & 86.07 \\ 
LA-Net & 20 & 87.12 & 61.72 & 86.85 \\ 
\textbf{NLA(Ours)}  & 20 & \textbf{87.60$\pm$0.13} & \textbf{63.25$\pm$0.04} & \textbf{87.64$\pm$0.2} \\ 
\midrule
Baseline & 30 & 75.50 & 52.16 & 79.77 \\ 
SCN & 30 & 77.46 & 55.05 & 82.47 \\ 
RUL & 30 & 82.06 & 56.93 & 83.90 \\ 
EAC & 30 & 84.42 & 58.91 & 85.44 \\ 
LA-Net & 30 & 85.33 & 60.82 & 86.01 \\ 
\textbf{NLA(Ours)}  & 30 & \textbf{86.71$\pm$0.16} & \textbf{62.48$\pm$0.14} & \textbf{86.97$\pm$0.04} \\ \bottomrule
\end{tabular}}
\caption{Comparison of overall accuracy with other methods under different noise ratios.}
\label{tab:noise_levels}
\end{table}

\paragraph{Different Noise Levels.}
We evaluate the robustness of NLA across three noise levels on the FERPlus, RAF-DB, and AffectNet datasets. Following previous studies \cite{zhang2021relative_RUL, zhang2022learn_EAC}, we corrupt 10\%, 20\%, and 30\% of the training labels by randomly flipping them to other categories. Experiments are conducted using five different random seeds, and we report the mean and standard deviation of the overall accuracy.
As shown in Table~\ref{tab:noise_levels}, NLA consistently outperforms other methods across all noise levels, achieving significant improvements over the baseline with gains ranging from 5.38\% to 10.01\%. This highlights the effectiveness of our method in handling noisy labels, even under extreme noise levels, through noise-aware adaptive weighting. Furthermore, compared to the state-of-the-art LA-Net, our method achieves average improvements of 0.93\%, 1.57\%, and 0.83\% on RAF-DB, AffectNet, and FERPlus, respectively. Considering that LA-Net employs an additional landmark-based backbone, the gains from our single-backbone model are particularly notable.

\begin{table}[t!]
\centering
\resizebox{0.95\linewidth}{!}{
\begin{tabular}{cccc}
\toprule
\textbf{Method} & \textbf{Imbalance} & \textbf{Overall} & \textbf{Mean} \\ \midrule
Baseline & 50 & 83.28 & 64.69 \\ 
BBN & 50 & 85.01 & 71.57 \\ 
EAC & 50 & 87.09 & 73.02 \\ 
MEK & 50 & 87.65 & 77.11 \\ 
\textbf{NLA(Ours)} & 50 & \textbf{87.97$\pm$0.26} & \textbf{78.05$\pm$0.14} \\ 
\midrule
Baseline & 100 & 80.96 & 55.12 \\ 
BBN & 100 & 83.44 & 67.92 \\ 
EAC & 100 & 85.79 & 69.80 \\ 
MEK & 100 & 86.47 & 73.06 \\ 
\textbf{NLA(Ours)} & 100 & \textbf{87.98$\pm$0.30} & \textbf{73.34$\pm$0.51} \\ 
\midrule
Baseline & 150 & 80.11 & 56.53 \\ 
BBN & 150 & 82.92 & 65.49 \\ 
EAC & 150 & 84.13 & 68.66 \\ 
MEK & 150 & 85.20 & 70.33 \\ 
\textbf{NLA(Ours)} & 150 & \textbf{86.34$\pm$0.05} & \textbf{70.49$\pm$0.57} \\ \bottomrule
\end{tabular}}
\caption{Comparison with other methods under different imbalances on RAF-DB.}
\label{tab:imbalance}
\end{table}

\paragraph{Different Imbalance Factors.}
To examine robustness against severe class imbalance, we follow established methods in handling imbalance in image classification~\cite{cao2019learning_im1, cui2019class_im2} by creating varying degrees of imbalance in the RAF-DB dataset using factors of 50, 100, and 150. The imbalance factor is the ratio of the number of training samples in the largest class to the number in the smallest class. We also conduct experiments using five random seeds, reporting the mean and standard deviation of overall and mean accuracy.
Table~\ref{tab:imbalance} shows NLA's superior performance across all imbalance factors. 
Compared to MEK, the current state-of-the-art in handling imbalance, NLA outperforms it in both overall accuracy and mean accuracy. 
This indicates that NLA effectively discriminates between minority and majority classes without bias. 

\begin{table}[t]
\resizebox{\linewidth}{!}{
\begin{tabular}{cccccccc}
\toprule
\multicolumn{1}{c}{\multirow{2}{*}{\textbf{Settings}}} & \multicolumn{5}{c}{\textbf{Components}} & \multirow{2}{*}{\textbf{Overall}} & \multirow{2}{*}{\textbf{Mean}} \\
\multicolumn{1}{c}{}  & CE  & NAW-CE  & $\mathcal{L}_1$ & CAM & JSD &                          &                       \\ \midrule
(a)       &\checkmark   &           &    &     &             & 87.42   & 78.53   \\
(b)       &             &\checkmark &    &     &             & 88.17   & 81.40   \\
(c)       &\checkmark   &           &    &     & \checkmark  & 87.78   & 81.01   \\
(d)       &             &\checkmark &\checkmark  &     &     & 89.18   & 82.87   \\
(e)       &             &\checkmark &    &\checkmark   &     & 89.15   & 83.02   \\
(f)       &             &\checkmark &    &     &\checkmark   & \textbf{89.93} & \textbf{83.87}   \\ \bottomrule
\end{tabular}
}
\caption{Ablation study results on RAF-DB.}
\label{tab:ablation_new}
\end{table}

\begin{figure*}[t!]
    \centering
    \includegraphics[width=1\linewidth] 
    {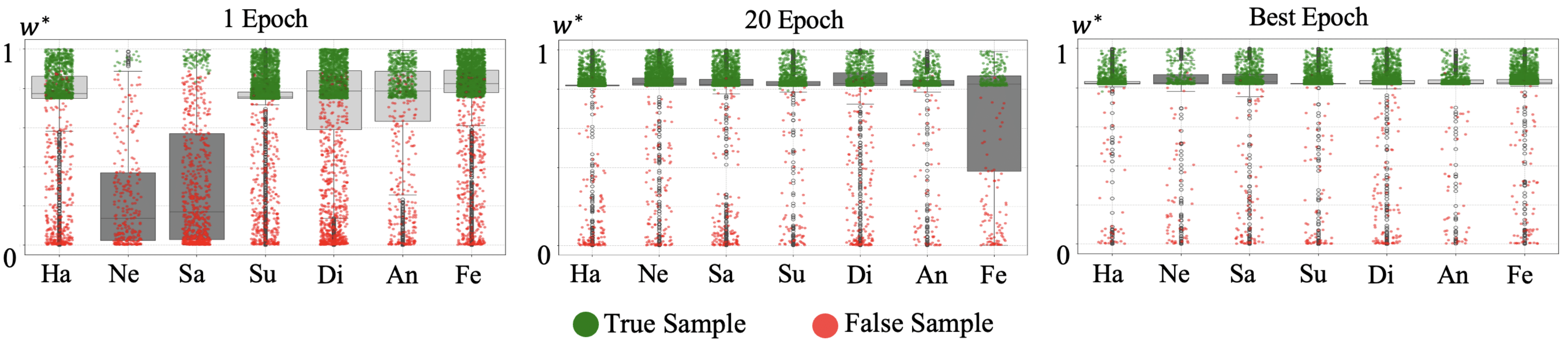} 
    \caption{\textbf{Visualization of the training process of our method.} This figure demonstrates how our method enhances discriminative ability by adaptively assigning weights to each sample through NAW. 
     }
    \label{fig:training_process_visualization} 
\end{figure*}

\subsection{Ablation Studies}
In Table~\ref{tab:ablation_new}, we present the results of an ablation study on the RAF-DB dataset, analyzing the contributions of each component within the NLA framework using a ResNet18 backbone. 
We compare the model using standard cross-entropy loss (CE) with the proposed NAW-based cross-entropy loss (NAW-CE). In setting (b), performance improves by 0.75\% in overall accuracy and 2.87\% in mean accuracy over the baseline (a), highlighting NAW’s effectiveness in adjusting sample weights.
Additionally, when regularization techniques are incorporated, as shown in settings (c) through (f), the model's performance improves even further. In these settings, $\mathcal{L}$1 represents the L1 regularization loss, and CAM refers to Class Activation Maps~\cite{zhou2016learning_CAM}.
Notably, setting (f), which corresponds to our method, achieves the highest performance with an overall accuracy of 89.93\% and a mean accuracy of 83.87\%. This significant enhancement underscores how Jensen-Shannon Divergence (JSD) amplifies the benefits of NAW by enforcing consistent distributional regularization.
We also evaluate the framework under conditions with 30\% noise and a class imbalance factor of 150. Replacing CE with NAW-CE leads to significant improvements—19.55\% in mean accuracy under noise and 7.97\% under class imbalance. 
These findings demonstrate that the proposed components of NLA significantly improve the model's robustness and reliability. 

\begin{figure}[t]
    \centerline{\includegraphics[width=1\linewidth]{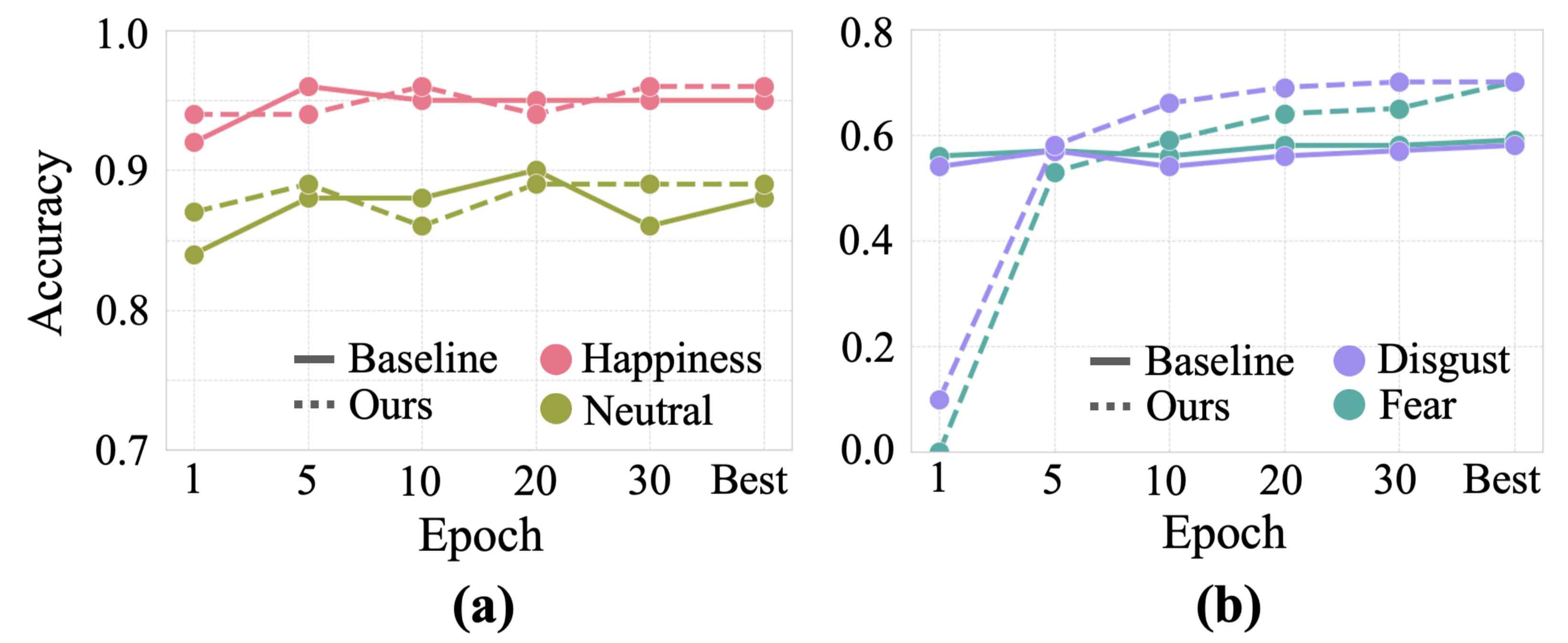}}
    \caption{Comparison of test accuracy across epochs for majority classes (a) and minority classes (b) on RAF-DB.}
    \label{fig:Accuracy_graph} 
\end{figure}

\subsection{Visualization Analysis} 
We visualize the training process of our method to analyze how the model develops class-specific discriminative ability.
In Fig.~\ref{fig:training_process_visualization}, the $y$-axis represents the noise-aware adaptive weight, $w^*$, assigned to each sample, and the x-axis shows the different classes (Ne: Neutral, Ha: Happiness, Sa: Sadness, Su: Surprise, An: Anger, Di: Disgust, Fe: Fear). The gray boxes indicate the Inter-Quartile Range, illustrating the sample distribution for each class. 
In the first epoch, samples in majority classes receive higher weights, while those in minority classes receive less attention.
However, as training progresses, our model gradually assigns higher weights to minority classes while maintaining the higher weights to the majority classes. 
Fig.~\ref{fig:Accuracy_graph} further supports this behavior, as (b) shows that our model initially exhibits low accuracy for minority classes but improves over time, whereas the baseline performance saturates.
These results demonstrate that the proposed model effectively handles class imbalance by adaptively adjusting weights based on sample ambiguity.

\begin{figure}[t!]
    \centerline{\includegraphics[width=1\linewidth]{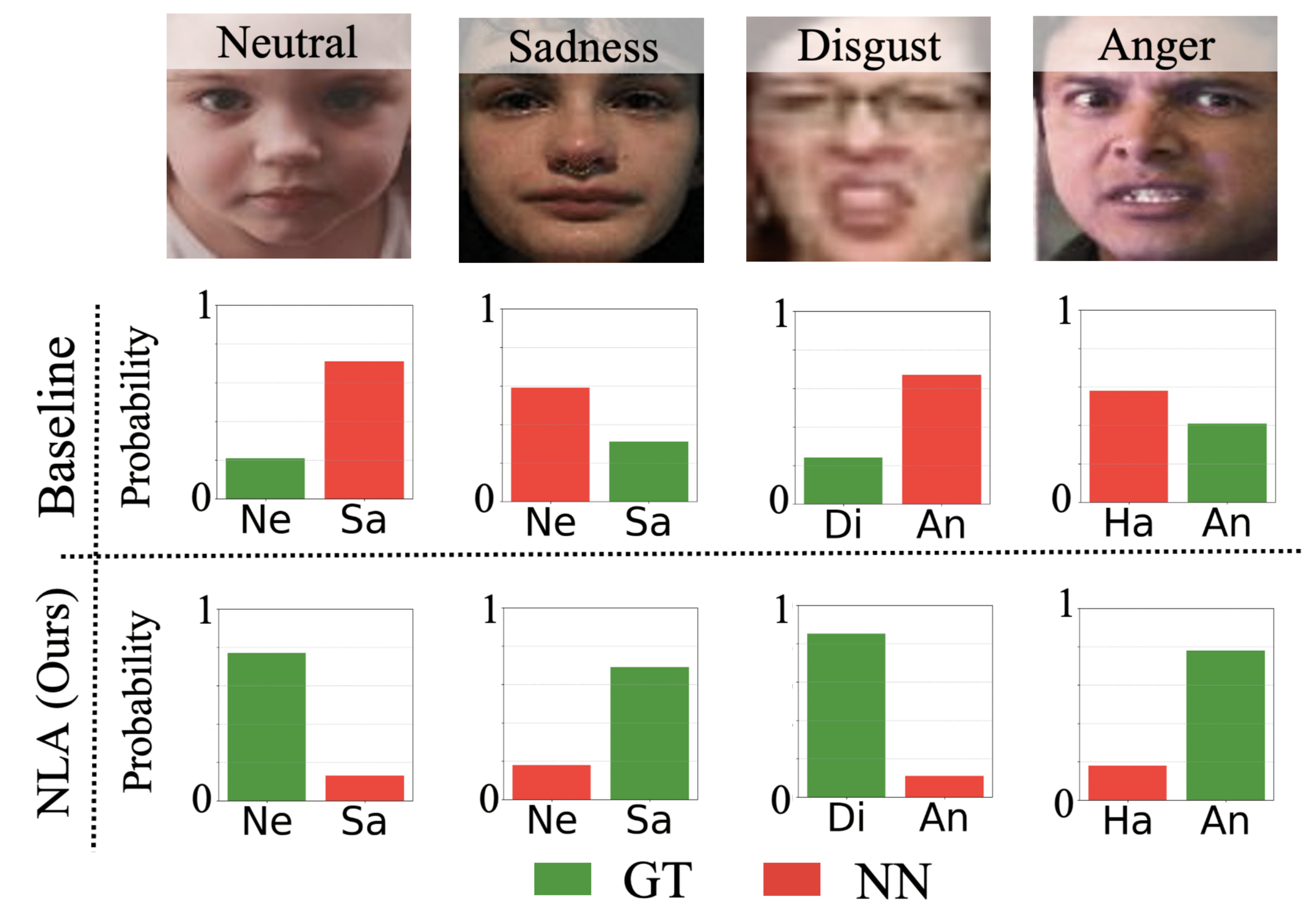}}
    \caption{\textbf{Comparison of results for ambiguous samples between the baseline and our model.}
    The baseline model confuses the prediction probabilities between the GT and NN, whereas our model correctly predicts by improving the prediction probability for the GT with a large margin.} 
    \label{fig:ambigous_samples} 
\end{figure}

Moreover, Fig.~\ref{fig:ambigous_samples} compares the baseline model and our NLA in handling ambiguous samples across different facial expression classes.
The top row shows ambiguous images from the RAF-DB dataset misclassified by the baseline model, with bar graphs below each image displaying predicted probabilities: green for GT and red for NN.
The baseline model often assigns similar probabilities to GT and NN, causing misclassification. In contrast, the bottom charts show that our model significantly increases the GT probability, creating a larger margin over the NN and resulting in correct classifications. This figure clearly demonstrates the effectiveness of NLA in accurately predicting ambiguous samples by enhancing the model's discriminative ability and reducing confusion with the NN.

\section{Conclusion}
In this paper, we propose a novel framework, Navigating Label Ambiguity (NLA), which addresses label ambiguity to mitigate both noise and class imbalance in facial expression recognition (FER). To the best of our knowledge, this is the first attempt to tackle both problems within a single framework. Our approach employs Noise-aware Adaptive Weighting (NAW) and consistency regularization to dynamically adjust weights based on sample ambiguity, enabling the model to focus on ambiguous samples while reducing the impact of noise. Extensive experiments demonstrate that NLA achieves superior overall and mean accuracy across multiple FER datasets, proving its robustness. Furthermore, while NLA is designed for FER, its ability to address label ambiguity in noisy or imbalanced data indicates promising applicability to a broader range of tasks, which we leave as an avenue for future work.

\section{Acknowledgments}
This research was supported by Field-oriented Technology Development Project for Customs Administration through National Research Foundation of Korea(NRF) funded by the Ministry of Science \& ICT and Korea Customs Service(2022M3I1A1095154), and supported by KIST Institutional Program (Project No. 2E33001).

\bigskip
\bibliography{aaai25}
\end{document}